%% file: main.tex
\documentclass[journal]{IEEEtran}
\usepackage{url}
\usepackage{graphicx}
\usepackage{amssymb}
\usepackage{amsmath}
\usepackage{amsthm}
\usepackage{booktabs}
\usepackage{algorithm}
\usepackage{algorithmic}
\usepackage{subfigure}
\usepackage{multirow}
\usepackage{graphicx}
\usepackage{color}
\usepackage{makecell}
\usepackage{subcaption}
\usepackage{tabularx}
\usepackage{listings}
\usepackage{sidecap}
\usepackage{balance}
\usepackage{pifont}
\usepackage[colorlinks,linkcolor=red,citecolor=blue]{hyperref}

\usepackage{colortbl}
\definecolor{cvprblue}{rgb}{0.21,0.49,0.74}

\usepackage[most]{tcolorbox}

\definecolor{dg}{rgb}{0,0.694,0.298}
\definecolor{purple}{rgb}{0.4,0.176,0.569}
\definecolor{Gray}{gray}{0.6}
\definecolor{royalblue}{RGB}{65,105,225}

\usepackage[capitalize]{cleveref}
\crefname{section}{Section}{Secs.}
\Crefname{section}{Section}{Sections}
\Crefname{table}{Table}{Tables}
\crefname{table}{Tab.}{Tabs.}

\PassOptionsToPackage{citecolor=royalblue,breaklinks,colorlinks,pagebackref}{hyperref}

\definecolor{royalblue}{RGB}{65,105,225} 
\usepackage{xspace}
\makeatletter
\DeclareRobustCommand\onedot{\futurelet\@let@token\@onedot}
\def\@onedot{\ifx\@let@token.\else.\null\fi\xspace}

\makeatother

\ifCLASSINFOpdf
\else
\fi
\begin{document}
%
\title{A Vision for Auto Research with LLM Agents}
%
%
%

\author{
Chengwei Liu\textsuperscript{1},
Chong Wang\textsuperscript{1},
Jiayue Cao\textsuperscript{2},
Jingquan Ge\textsuperscript{1},
Kun Wang\textsuperscript{1},
Lyuye Zhang\textsuperscript{1},
Ming-Ming Cheng\textsuperscript{2},
Penghai Zhao\textsuperscript{2},
Tianlin Li\textsuperscript{1},
Xiaojun Jia\textsuperscript{1},
Xiang Li\textsuperscript{2},
Xingshuai Li\textsuperscript{2},
Yang Liu\textsuperscript{1},
Yebo Feng\textsuperscript{1},
Yihao Huang\textsuperscript{1},
Yijia Xu\textsuperscript{1},
Yuqiang Sun\textsuperscript{1},
Zhenhong Zhou\textsuperscript{1},
Zhengzi Xu\textsuperscript{1}\\
\textsuperscript{1}Nanyang Technological University, 
\textsuperscript{2}Nankai University
\thanks{Authors are listed in alphabetical order.}
}


%
%

\markboth{Journal of \LaTeX\ Class Files,~Vol.~14, No.~8, August~2015}%
{Shell \MakeLowercase{\textit{et al.}}: Bare Demo of IEEEtran.cls for IEEE Journals}
%



\maketitle

\begin{abstract}
This paper introduces Agent-Based Auto Research, a structured multi-agent framework designed to automate, coordinate, and optimize the full lifecycle of scientific research. Leveraging the capabilities of large language models (LLMs) and modular agent collaboration, the system spans all major research phases, including literature review, ideation, methodology planning, experimentation, paper writing, peer review response, and dissemination. By addressing issues such as fragmented workflows, uneven methodological expertise, and cognitive overload, the framework offers a systematic and scalable approach to scientific inquiry. Preliminary explorations demonstrate the feasibility and potential of Auto Research as a promising paradigm for self-improving, AI-driven research processes.
\end{abstract}


%
\IEEEpeerreviewmaketitle

\section{Introduction}
\IEEEPARstart{S}{cientific} research is undergoing a profound transformation driven by the emergence of automated machine learning (AutoML) systems, large language models (LLMs), and multi-agent collaboration frameworks \cite{IEEEexample:bluebookstandard}. These technologies have begun to redefine the boundaries of computational creativity, enabling artificial intelligence (AI) systems to perform increasingly complex cognitive tasks. As AI continues to scale, both in terms of model capacity and access to domain knowledge, its role in shaping the future of research is no longer speculative but inevitable. These advances point toward an emerging paradigm we refer to as ``Auto Research'', which is a structured multi-agent framework for automating and enhancing the full spectrum of scientific investigation.

Despite the growing availability of digital tools and scientific literature, researchers continue to face persistent obstacles that hinder progress. Methodological knowledge remains unevenly distributed, making it difficult for individuals to identify effective approaches or determine the feasibility of novel ideas without extensive trial and error. The research process itself has become increasingly fragmented—spanning literature review, hypothesis formulation, experimental design, result analysis, and manuscript writing, etc.—often requiring expertise across multiple domains. This fragmentation not only leads to inefficiencies but also places heavy demands on individual researchers to master diverse skill sets. At the same time, the human capital underpinning research is highly variable: expertise is difficult to scale, team composition often lacks stability, and supervisory resources are scarce or uneven. These challenges are particularly acute for students and early-career researchers, who frequently lack consistent methodological guidance and structured feedback. Compounding these issues is the limited systematization of research reasoning. Unlike engineering disciplines that benefit from modular design and reusable components, scientific problem solving remains largely ad hoc and intuition-driven. As a result, critical decisions (such as choosing appropriate methods or assessing the significance of a research problem) lack transparency and reproducibility, making the research process both labor-intensive and cognitively opaque.

These challenges call for a systematic rethinking of the research process, not merely through isolated tools but through an integrated architecture that embodies reasoning, coordination, and adaptability. Recent advances in large language models and multi-agent frameworks offer a compelling foundation for this transformation. LLM-based agents are not only capable of understanding and generating scientific content, but also of participating in decision-making, decomposing complex tasks, and integrating domain feedback. Their flexibility allows them to operate across a wide range of research domains, while their composability enables the design of workflows that are modular, extensible, and context-aware. These properties make them particularly well-suited to address the fragmentation of workflows, variability in human expertise, and cognitive burden identified earlier.

Building on these capabilities, we propose Agent-Based Auto Research as a structured research system designed to automate, coordinate, and optimize the full lifecycle of scientific investigation. The research pipeline is conceptualized as a sequence of distinct yet interdependent phases, each supported by specialized agents operating within structured workflows: \ding{172} Literature. The agents automate the literature review process by synthesizing and analyzing existing research, identifying gaps, and guiding future directions. They assist in topic refinement, paper retrieval, and keyword generation to streamline literature research. \ding{173} Idea. The agents identify existing research problems and propose new solutions by generating novel algorithms, models, and techniques. They also uncover new research problems by analyzing literature and real-world needs, enabling the exploration of uncharted research areas. \ding{174} Method. The Method Planner breaks down complex research problems into manageable tasks and generates high-level plans, while the Heuristic Solution Designer autonomously selects suitable methods using heuristic evaluations to ensure efficient execution of the research plan. \ding{175} Experiment. The agents assist in defining experimental setups by identifying benchmarks, establishing baselines, and selecting metrics. They also generate executable code based on the methodology and analyze experimental results to extract meaningful insights and create visualizations. \ding{176} Paper. The agents streamline the paper writing process by generating drafts for various sections, including the abstract, introduction, methodology, and evaluation. They ensure proper structure, maintain logical consistency, and help in synthesizing related work, limitations, and future directions. \ding{177} Evaluation. The multi-agent systems evaluate academic papers by simulating peer-review dynamics. Agents assess papers based on novelty, rigor, relevance, verifiability, and presentation using a structured evaluation workflow, integrating Chain-of-Thought reasoning and dynamic processes for continuous refinement. \ding{178} Rebuttal. The agents facilitate rebuttal writing by extracting and classifying reviewer feedback, prioritizing critical comments, and generating structured, concise responses. This ensures major concerns are addressed effectively, maintaining professionalism and clarity. \ding{179} Promotion. The agents optimize research promotion strategies by tailoring content to different paper types and platforms. They continuously refine strategies based on real-time engagement data, using specialized agents to retrieve papers, summarize content, and generate targeted promotional material. To assess the practicality of this conceptual framework, preliminary explorations were conducted across selected phases, demonstrating the feasibility and promise of Auto Research as a structured, self-improving, agent-driven model of scientific inquiry. 

Auto research is not merely a toolchain or automation layer—it represents a shift toward a new epistemological model of doing science. By treating research as a modular, interpretable, and improvable process, Auto Research has the potential to democratize scientific inquiry, mitigate human limitations, and accelerate methodological innovation across disciplines. As large-scale foundation models and collaborative agent frameworks continue to evolve, we believe Auto Research offers a forward-looking vision for the co-evolution of human and machine intelligence in scientific discovery. To further articulate this vision, the remainder of the paper is organized as follows. We first examine the foundational insights from scaling laws in AI, which motivate the feasibility of automating scientific processes at scale. We then introduce the design of our multi-agent Auto Research framework, followed by a breakdown of its key modules across the full research lifecycle. Next, we present a series of exploratory studies to assess the practicality and adaptability of the proposed architecture. Finally, we reflect on the broader implications of this paradigm, discuss the distinction between cumulative and disruptive research, the importance of a meta-method for adapting AI to various methodologies, and how AI facilitates knowledge creation by integrating diverse sources to accelerate scientific discovery.

\section{Scaling Law in AI}
The concept of scaling laws in AI describes how increasing key resources, such as model parameters, training data, and computational power predictably enhances performance. Empirical studies have demonstrated that deep learning models, particularly in language processing and vision, exhibit power-law relationships where loss reduction and capability growth follow consistent trends as scale increases. This phenomenon suggests that larger models trained on more data tend to generalize better, demonstrating emergent abilities such as in-context learning and zero-shot reasoning. Scaling laws reveal that AI performance improvements are not merely incremental but follow predictable mathematical patterns.  Early work in this area, notably from OpenAI and DeepMind, has shown that model loss declines smoothly as a power function of computational budget and parameter count.

Beyond AI model performance, scaling laws offer insights into the automation of research itself.  If research workflows, including literature review, hypothesis generation, and experiment design, can be systematically scaled using AI agents, similar power-law behaviors may emerge.  The application of multi-agent systems, where specialized AI components collaborate, suggests that automated research may exhibit its own scaling properties, optimizing efficiency through structured scaling of computation, knowledge aggregation, and decision-making.  The potential of such a system extends beyond efficiency gains, as interactions between AI-driven research agents may unlock new forms of interdisciplinary synthesis, accelerating innovation in unexpected ways.

Understanding scaling laws in AI provides a foundation for exploring how research automation can be structured.  By identifying the key variables that drive progress, automated scientific discovery could potentially follow predictable trends, accelerating the generation of new knowledge in a systematic and scalable manner.  As AI systems evolve, deriving scaling laws for automated research may provide a framework for designing future scientific exploration, ensuring that computational and methodological resources are allocated optimally to maximize discovery.

\section{Multi-Agent System-Based Research}
\begin{figure*}[tb]
\centering
\includegraphics[width=0.95\linewidth]{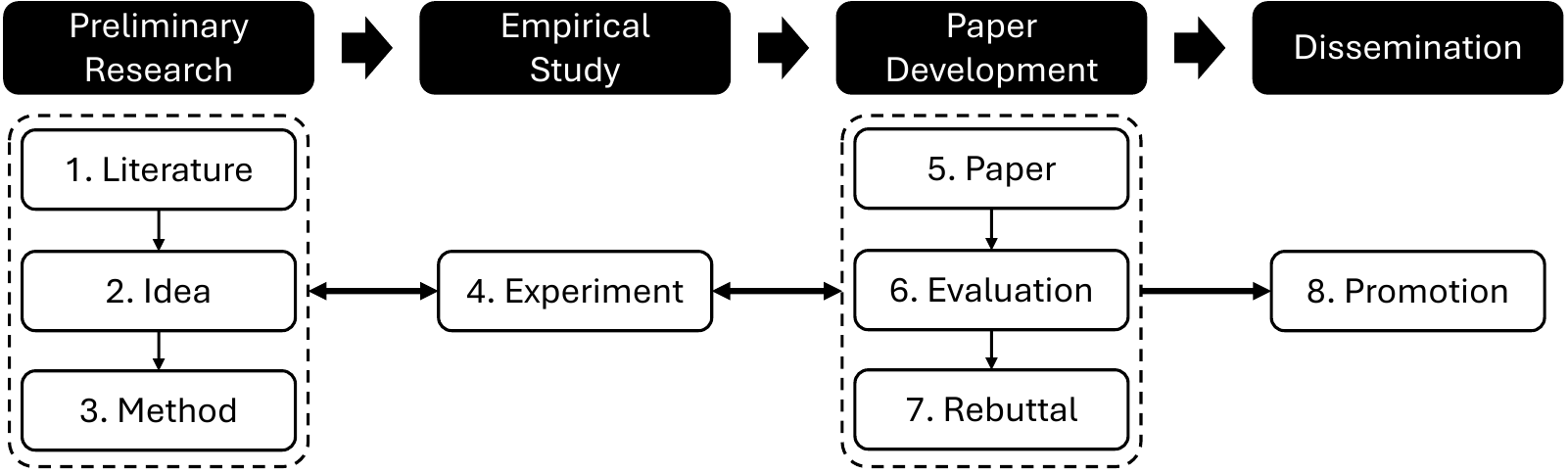}
\caption{The pipeline of our agent-based auto research framework.}
\label{fig:main_pipeline}
\end{figure*}

This Agent-Based Auto Research Framework (see Figure~\ref{fig:main_pipeline}) is designed to streamline and automate the research workflow, facilitating a structured and iterative approach to academic studies. The framework consists of four key stages:

\subsubsection{Preliminary Research}
This phase involves an iterative process of literature review, idea generation, and method development, forming the conceptual foundation for the study.

\subsubsection{Empirical Study}
The developed method is validated through experimentation, which interacts dynamically with the preliminary research phase for continuous refinement.

\subsubsection{Paper Development}
Once empirical results are obtained, the framework supports paper writing, self-evaluation, and rebuttal preparation, ensuring the manuscript meets academic standards and addresses peer review feedback.

\subsubsection{Dissemination}
Finally, the research findings are promoted through various channels to maximize impact and visibility.

By integrating automated agents at each stage, this framework enhances efficiency, reduces manual workload, and fosters a more structured research process. The detailed design of each part is presented in the following sections.

\subsection{Literature}
\label{sec:literature}

\subsubsection{Literature Review}

A literature review systematically synthesizes existing research and critically analyzes key findings, laying a solid foundation for identifying clear and meaningful research directions. Depending on their purposes and methodologies, literature reviews are commonly classified into various types, including narrative review, systematic review, and scoping review, \textit{etc.}~\cite{grant2009typology,ralph2022paving,zhao2024ror}. Typically, narrative reviews summarize topics broadly, systematic reviews rigorously evaluate research evidence based on explicit criteria, and scoping reviews identify the scope, range, and key characteristics of existing research. Despite their methodological importance, literature reviews are often costly to produce in terms of time, expertise, and cognitive effort. As the volume of academic publications continues to grow, the question of how to automate the literature review process—while maintaining quality and relevance—has become an increasingly active area of research.

Automating literature reviews~\cite{wang2024autosurvey,yan2025surveyforge,liang2025surveyx} typically involves structured workflows consisting of three critical stages: knowledge retrieval, content synthesis, and report generation. The first stage, knowledge retrieval, aggregates information from diverse sources including academic publications, preprints, blogs, technical reports, and informal online discussions. Due to the varied reliability of these sources, verifying the accuracy and credibility of information becomes an essential task. The second stage, content synthesis, involves systematically organizing the retrieved knowledge into structured frameworks tailored to specific research objectives, thus providing a solid foundation for subsequent in-depth literature research. Lastly, the report generation stage converts these structured insights into clear and accessible formats, producing narratives or structured outputs that align with both human and AI agent usage.

In general, literature reviews in automated research go beyond conventional summaries by systematically identifying a range of potential research directions related to specific practical problems. They function similarly to structured mapping studies~\cite{petersen2008systematic}, carefully examining extensive literature to outline the various methods researchers have used to tackle real-world issues. By categorizing existing studies in this systematic way, automated literature reviews help AI agents effectively determine which research directions appear most promising and feasible. After establishing these clear directions from the literature review, the agents naturally proceed to the next step, performing more targeted and detailed literature research.

\subsubsection{Related Work Review}

After reading the literature review and confirming the research direction, the next indispensable stage is an in-depth related work review. This process serves three main purposes: (1) to sort out existing related research results, (2) to analyze their advantages, disadvantages, and applicable scenarios, and (3) to clarify the positioning and distinctions of the proposed research work.

The overall process of related work review is divided into four steps:

\begin{enumerate}
    \item \textbf{Clarify research direction and technical keywords.} Define the research scope clearly and concisely using appropriate technical terms.
    
    \item \textbf{List existing work or tools.} Prepare a detailed list of relevant papers or tools you want to cover. Since large models may not be aware of the latest literature you’ve read, manually identifying key items is essential.
    
    \item \textbf{Generate initial draft using prompts.} Compose a prompt and use the large model to generate a first draft of the related work section.
    
    \item \textbf{Manual polishing and citation completion.} After generation, manually add precise citations (e.g., BibTeX entries), adjust the writing style to maintain consistency with the paper, and refine the final sentence of each paragraph to reflect your research emphasis.
\end{enumerate}

\subsection{Idea}

There are several types of ideas for research papers. For each type, an AI agent can be designed to generate corresponding ideas.

\subsubsection{Existing Problems with New Solutions}

A large number of research papers are published to solve existing problems with new solutions. These solutions can take the form of new algorithms, models, architectures, or techniques.

\subsubsection*{Problem Decomposition}
For complex problems, it is crucial to break down a high-level research challenge into tractable sub-problems. By prompting an agent with a broad research area and asking it to identify implicit dimensions, hidden assumptions, or orthogonal aspects based on literature and expert practices, the agent can suggest plausible ways to partition the problem space.

For example, in vulnerability detection tasks, it may analyze the literature and recognize that certain vulnerabilities can be categorized by their preconditions, effects, or environmental assumptions. These structured decompositions clarify the scope and enable focused investigations into individual components—potentially inspiring new detection techniques, benchmarks, or formal models.

\subsubsection*{Problem Generalization}
LLM-based agents can facilitate problem generalization by identifying latent commonalities across seemingly disparate research problems and proposing unifying abstractions. By analyzing diverse studies, agents can detect structural patterns—such as data dependencies or threat models—that underlie various methods. This enables generalization and development of reusable frameworks while uncovering connections between subfields.

\subsubsection*{Directly Using New Techniques}
Advances in technology often enable direct application of new techniques to existing problems. LLM-based agents can help identify opportunities where recent innovations—such as new architectures or learning paradigms—may address long-standing challenges. This involves reasoning about technological capabilities and simulating analogies across domains to guide re-contextualization and application.

\subsubsection*{Combination of Existing Techniques}
When technological breakthroughs are unavailable, researchers can combine existing techniques for better performance. LLM-based agents can synthesize hybrid solutions by leveraging diverse methodological knowledge, such as static analysis, graph learning, or symbolic execution. The agent reasons about complementary strengths and may suggest integrations or ensemble approaches that balance accuracy, scalability, and robustness.

\subsubsection{New Problems}

In addition to solving existing problems with new solutions, researchers can also identify new problems that are not yet well addressed.

\subsubsection*{Re-challenging Existing Solutions}
Existing solutions may prove ineffective or non-robust in new scenarios. LLM-based agents can systematically examine the assumptions and constraints of current methods to uncover weaknesses. By simulating edge cases or shifts in settings, agents can generate counterexamples or alternative evaluations that expose brittleness—especially in dynamic or rapidly evolving domains.

\subsubsection*{Finding New Research Fields}
LLM-based agents can help discover emerging or underexplored research fields. This can involve identifying real-world needs, synthesizing new requirements from existing problem reviews, or proposing novel targets. By processing literature, documentation, and online discourse, agents can detect trends that haven’t yet formed formal problems and propose interdisciplinary opportunities that link disparate domains.

\subsubsection*{Empirical Studies}
Empirical studies can re-assess existing solutions to uncover new problems or recurring limitations. LLM-based agents can scale empirical research by gathering and analyzing large datasets—e.g., from repositories, issue trackers, or papers. They can replicate experiments, extract metrics, and code qualitative data (e.g., GitHub issues) to surface overlooked pain points and support comprehensive empirical evaluations.

\subsubsection*{Survey Paper Generation}
Survey papers summarize existing work in a field. LLM-based agents are well-suited for this task due to their ability to synthesize and summarize large volumes of information. They can classify methods, compare approaches, and highlight strengths and limitations—facilitating comprehensive, high-quality surveys.

\input{method}

\subsection{Experiment}

Experiments conducted during the research process aim to verify the feasibility and effectiveness of the proposed methodological design. To achieve this purpose, the experimental phase typically involves three main components: establishing the experimental setup, implementing the methodology, and analyzing the results. This section introduces each component in sequence.

\subsubsection{Experimental Setup}

LLM-based agents can play a pivotal role in designing experimental setups by systematically determining benchmarks, baselines, metrics, and models. The setup design typically involves the following aspects:

\begin{itemize}
    \item \textbf{Specification of Research Objectives and Constraints:} Clearly define the problem to be addressed, expected outcomes, and constraints. Agents can analyze relevant datasets and prior knowledge to support this process.
    
    \item \textbf{Benchmark Identification:} Agents use data mining to extract benchmarks from existing literature and databases. They perform trend and comparative analyses to ensure relevance and alignment with current standards.
    
    \item \textbf{Baseline Establishment:} By reviewing prior experiments and applying statistical modeling, agents can define baseline performance metrics. Hypothesis testing may be used to simulate different baseline scenarios.
    
    \item \textbf{Metric Selection:} Agents prioritize metrics based on their alignment with research goals. They recommend metrics that are sensitive, specific, and adaptable to evolving objectives or incoming data.
    
    \item \textbf{Model Selection and Configuration:} Agents propose candidate models from machine learning, statistics, or simulation. Configuration is optimized via grid search, random search, or Bayesian techniques. Simulation ensures alignment with expectations.
    
    \item \textbf{Automation and Iteration:} The experimental design process is iterative. Agents learn from feedback and continuously refine benchmarks, baselines, and metrics. Scalability is maintained even as complexity increases.
\end{itemize}

\subsubsection{Methodology Implementation}

LLM-based agents can interpret a methodology specification and generate executable code. This process includes:

\begin{itemize}
    \item \textbf{Implementation Intention Identification:} Agents extract research goals, techniques, and expected outputs from researcher input. They determine whether the methodology pertains to analysis, simulation, hypothesis testing, or model development.
    
    \item \textbf{Code Implementation:} Agents generate syntactically correct and semantically appropriate code. They test and debug the output to ensure functionality, applying optimization techniques to improve performance and scalability. The process remains flexible to adapt to feedback or changes in methodology.
    
    \item \textbf{Integration:} Code is integrated into the research workflow. Agents ensure compatibility with existing systems and monitor execution, providing feedback, alerts, and real-time metrics for performance and correctness.
\end{itemize}

\subsubsection{Analysis of Experimental Results}

LLM-based agents can automate the interpretation of experimental outcomes using advanced data analysis capabilities. This includes:

\begin{itemize}
    \item \textbf{Data Processing and Interpretation:} Agents ingest raw data (structured or unstructured), clean and normalize it, and identify key trends or anomalies aligned with research goals.
    
    \item \textbf{Pattern Recognition and Insight Extraction:} Agents use ML techniques to detect correlations and trends. This supports discovery of meaningful insights and helps refine hypotheses.
    
    \item \textbf{Automated Reporting and Visualization:} Agents generate comprehensive reports, including statistical summaries, visualizations, and narrative explanations. Charts and graphs enhance interpretability.
    
    \item \textbf{Continuous Feedback and Iterative Analysis:} Agents refine their models and interpretations based on new data and researcher feedback. Real-time responsiveness ensures ongoing relevance.
    
    \item \textbf{Decision Support and Predictive Analysis:} Agents can simulate future outcomes based on current trends, supporting informed decisions about subsequent experiments or modifications to the current approach.
\end{itemize}

Overall, LLM-based agents streamline experimentation by aligning setups with research objectives, generating implementation code, and analyzing results. Their ability to operate iteratively and at scale enhances efficiency, rigor, and insight quality across the empirical research cycle.

\subsection{Paper}
\input{paper_writing}

\subsection{Evaluation}

\subsubsection{Overview}
This system employs a multi-agent architecture where specialized roles, such as analysts, critics, validators, and moderators, collaboratively evaluate academic papers through structured debates and iterative refinement. By integrating real-time search engines (e.g., Google Scholar, Semantic Scholar) for evidence retrieval and avoiding predefined databases, the framework ensures evaluations reflect the latest research trends. Agents simulate peer-review dynamics, using Chain-of-Thought (CoT) reasoning to dissect each criterion (novelty, rigor, relevance, verifiability, and presentation) and reach consensus-driven conclusions. The process emphasizes transparency, adaptability, and scalability while minimizing biases inherent to static datasets or single-reviewer assessments.

\subsubsection{Agent Roles and Collaborative Dynamics}
Three major specialized roles are designed as agents to collaborate to derive academic paper evaluation.

\paragraph{Analysts}
Initiate evaluations by dissecting the paper’s structure, extracting core claims, methodologies, and results. They generate preliminary assessments using summarization techniques and keyword-driven search queries.

\paragraph{Critics}
Challenge assessments by retrieving contradictory or overlapping studies via search APIs, posing questions like, “Does this approach address limitations cited in recent reviews?” or “How does this differ from method X in [retrieved paper]?” Critics employ counterfactual reasoning to stress-test assumptions.

\paragraph{Validators}
Enforce domain-specific standards, verifying statistical methods, ethical compliance, and reproducibility. They cross-check data availability statements against repositories like GitHub and validate formatting against style guides (APA, IEEE).

\paragraph{Moderator}
Synthesizes inputs, resolves disputes, and assigns scores, ensuring final evaluations balance rigor with fairness.

\subsubsection{Evaluation Workflow and Chain-of-Thought Reasoning}
For each evaluation criterion, agents engage in a three-phase workflow:

\paragraph{1) Debate Phase}
Analysts propose initial scores based on extracted content (e.g., “This paper introduces a hybrid algorithm for gene sequencing”). Critics counter with evidence from dynamically retrieved papers. Validators intervene to flag methodological oversights or compliance gaps.

\paragraph{2) Refinement Phase}
Agents iteratively revise scores using CoT prompts. Analysts adjust justifications based on counterarguments. Validators reassess technical soundness.

\paragraph{3) Consensus Phase}
The moderator aggregates perspectives, applying weighted voting or conflict-resolution rules. If disagreement persists, validators' input is prioritized on technical uniqueness.

\subsubsection{Dynamic Evaluation Processes}
The framework evaluates academic papers across five core criteria widely accepted in the Software Engineering community~\cite{ICSE2026review}.

\paragraph{Novelty: Beyond Incremental Progress}
The system parses core contributions and hypotheses, querying scholarly databases for overlapping methods. Embedding models (e.g., SPECTER) quantify conceptual similarity. Iterative debates distinguish textual from conceptual novelty. Final scores emphasize underexplored gaps and methodological uniqueness.

\paragraph{Rigor: Methodological and Statistical Soundness}
Methodology sections are parsed to extract statistical tests and data pipelines. Validators flag concerns like p-hacking or missing control groups. Code repositories are executed in sandboxed environments to verify reproducibility. Cross-disciplinary standards (e.g., COREQ) ensure contextual fairness.

\paragraph{Relevance}
Topic modeling algorithms compare paper content with target venue scope. Critics examine whether contributions align with framing. The system may suggest alternative venues when scope mismatch arises, enhancing relevance.

\paragraph{Verifiability \& Transparency: Reproducibility as a Standard}
The system audits data/code repositories for FAIR compliance. Validators inspect metadata, IRB approvals, and ethical disclosures. Feedback includes actionable suggestions (e.g., uploading to Zenodo).

\paragraph{Presentation: Clarity and Scholarly Communication}
The system evaluates writing clarity, structure, and visual elements. Readability tools assess grammar and tone. CV-based models review figure quality and accessibility. Trade-offs between brevity and completeness are debated contextually.

\paragraph{Conflict Resolution and Consensus Building}
Conflicting evaluations (e.g., high novelty but low rigor) are resolved through weighted rules. Validators' assessments are prioritized over stylistic concerns. All debates are archived for transparency and traceability.

\subsection{Rebuttal}
To generate an effective rebuttal for software engineering conference reviews, we preprocess the review comments and classify them based on their sentiment and references to different sections of the submitted paper. This step ensures that we can systematically address concerns, reinforce contributions, and clarify ambiguous points.

\subsubsection{Text Extraction and Classification using LLMs}
To systematically process reviewer feedback, we employ a multi-step approach leveraging LLMs for structured text extraction and classification. The process iterates over each review, extracts individual feedback points, and classifies them based on three orthogonal aspects: their reference to the paper, comment type, and sentiment.

\paragraph{Step 1: Review Processing and Segmentation}
Given three reviewer comments in unstructured text format, we first convert them into a machine-readable format. The preprocessing pipeline includes:
\begin{enumerate}[leftmargin=10pt]
    \item \textbf{Text Conversion:} Extract review content using Optical Character Recognition (OCR) if necessary (e.g., for scanned PDFs).
    \item \textbf{Sentence Segmentation and Tokenization:} Apply basic NLP techniques to split text into individual sentences.
    \item \textbf{Paragraph Splitting:} Identify and separate paragraphs containing distinct feedback points.
\end{enumerate}

\paragraph{Step 2: Classification of Review Comments}
For each extracted feedback point, we classify the comment along three independent dimensions using LLM-based text classification:

\begin{itemize}
    \item \textbf{Linking Feedback to Paper Sections:} Each feedback point is mapped to one or more relevant sections of the paper, categorized as: Abstract, Introduction, Background, Methodology, Dataset, Experiment, Threats to Validity, and Conclusion.

    \item \textbf{Classifying the Type of Comment:}  
    Each feedback point is categorized into one of the following types:  
    \textbf{General Description}, \textbf{Strength}, \textbf{Weakness}, \textbf{Questions}, \textbf{Presentation}, \textbf{Rigor}, \textbf{Relevance}, \textbf{Novelty}, and \textbf{Reproducibility}.

    \item \textbf{Sentiment Analysis:}  
    We classify the sentiment using an Aspect-Based Sentiment Analysis (ABSA) approach~\cite{rusnachenko2024largelanguagemodelstargeted,simmering2023largelanguagemodelsaspectbased,zhou2024comprehensiveevaluationlargelanguage}, with the following labels:  
    \textbf{Positive}, \textbf{Negative}, and \textbf{Neutral}.
\end{itemize}

\paragraph{Step 3: Classification and Structured Output}
To automate classification, we design targeted prompts for an LLM (e.g., GPT-4, Claude, Gemini). The system processes each review segment individually, applying three classification prompts sequentially. An example is shown in Table~\ref{tab:classification_categories}.

\begin{table}[h]
    \centering
    \begin{tabular}{|c|c|c|}
        \hline
        \textbf{Linked Paper Section} & \textbf{Comment Type} & \textbf{Sentiment} \\ \hline
        Abstract & General Description & Positive \\ \hline
        Introduction & Strength & Negative \\ \hline
        Background & Weakness & Neutral \\ \hline
        Methodology & Question & \\ \hline
        Dataset & Presentation & \\ \hline
        Experiment & Rigor & \\ \hline
        Threats to Validity & Relevance & \\ \hline
        Implications and Discussion & Novelty & \\ \hline
        Conclusion & Reproducibility & \\ \hline
    \end{tabular}
    \caption{Enumeration of Classification Categories for Review Comments}
    \label{tab:classification_categories}
\end{table}

\subsubsection{Prioritization for Rebuttal Writing}
Once review comments are structured, we rank them for rebuttal writing. The ranking process follows a multi-factor prioritization strategy:

\begin{itemize}
    \item \textbf{Comment Type Importance:}  
    \textit{Weakness} or \textit{Question} comments are ranked higher than \textit{General Description} or \textit{Strength}. Presentation and reproducibility concerns are assigned medium priority.

    \item \textbf{Sentiment-Based Weighting:}  
    Negative comments receive the highest priority. Neutral feedback comes next, followed by positive comments (to reinforce strengths).

    \item \textbf{Paper Section Relevance:}  
    Comments on Methodology, Experiment, and Threats to Validity are prioritized over Presentation or Background.
\end{itemize}

Ranking scores are computed by LLMs based on the above dimensions.

\subsubsection{Response Generation with Word Count Optimization}
To ensure concise and persuasive rebuttals under word limits, we adopt a \textbf{multi-agent LLM framework} consisting of the following steps:

\begin{enumerate}[leftmargin=10pt]
    \item \textbf{Keyword Extraction for Concise Representation:}  
    A keyword agent identifies critical terms in feedback segments to reduce redundancy.

    \item \textbf{Initial Response Drafting:}  
    A generation agent writes polite and constructive responses for each ranked segment, including clarifications and supporting evidence.

    \item \textbf{Conciseness Optimization via LLM Compression Agent:}  
    A second agent compresses the draft responses to maximize clarity and persuasiveness within space constraints.

    \item \textbf{Response Allocation Based on Ranking:}  
    Word count is allocated based on ranking. Higher-ranked segments receive more detailed responses.

    \item \textbf{Word Count Calculation:}  
    A word counter ensures total length adheres to the rebuttal limit.

    \item \textbf{Coordinator Agent for Strategic Condensation:}  
    A meta-agent dynamically adjusts compression intensity for lower-priority segments to preserve detail for high-priority responses.

    \item \textbf{Iterative Refinement Until Word Limit is Met:}  
    If the total rebuttal exceeds the word limit, agents apply stronger compression strategies iteratively until constraints are met.
\end{enumerate}

This multi-agent LLM strategy ensures the rebuttal remains precise, persuasive, and balanced under strict length limits.

\subsection{Promotion}

\subsubsection{Research Questions}

\paragraph{RQ1: Diversified Promotion Generation Strategies}  
Currently, research papers are published in vast quantities, and they exhibit significant differences in type, such as trend-driven studies, technical reports, empirical studies, and rigorous theoretical analyses. When promoting these papers, distinct strategies are required.

Trend-driven papers emphasize dissemination and accessibility in promotion, necessitating engaging titles and content with a low reading barrier. Technical reports, in contrast, typically do not prioritize attention-grabbing titles but instead highlight core contributions at the beginning and end while presenting diverse experimental data and conclusions in the main body. Empirical studies derive conclusions through experimental methods, so their promotions should emphasize key findings in the title while providing intuitions and reasoning in the main content. For theoretical research, the promotion should maintain the integrity of the analysis or proof while using relatively simple content to assist comprehension.

To align promotion with the objectives of different types of papers, it must be sufficiently tailored to ensure effective dissemination that matches the papers' intended impact.

\paragraph{RQ2: Following Social Media Platform Preferences and Rules}  
When promoting research, various social media platforms allow users to post promotional articles or blogs. These platforms cater to different audiences, leading to significant differences in their recommendation algorithms and rules.

For English-language social media, \textit{Twitter (X)} favors concise, text-heavy posts with minimal imagery, requiring promotions to be engaging within a short format. \textit{Medium} prefers long-form, in-depth analyses that encourage extended reading time and interaction. \textit{Reddit} has fewer restrictions but relies on sparking discussions to drive engagement, as posts with active comment sections are more likely to be upvoted and remain visible within relevant topics.

For Chinese social media, \textit{WeChat Official Accounts} attract readers through appealing titles and preview cards, with metrics such as reading completion rate, likes, and favorites significantly impacting recommendation weight. \textit{Xiaohongshu (RedNote)} mandates an overview figure, which serves as the primary hook for readers. \textit{Zhihu} favors long-form content driven by a question-based format, where selecting the right question for submission can yield substantial exposure. \textit{Weibo} functions similarly to \textit{Twitter (X)}, emphasizing short, engaging posts.

Additionally, different social media platforms have varying levels of censorship requirements, with Chinese platforms generally enforcing stricter content regulations than their English counterparts. Some platforms impose unique posting rules, and many social media platforms disfavor external links, either reducing the visibility of posts that include them or outright blocking such posts—RedNote, for instance, actively blocks posts containing external links.

Given these differences, effective promotion must be platform-specific, ensuring content aligns with each platform's preferences and rules while avoiding pitfalls that could lead to reduced visibility or ineffective outreach.

\paragraph{RQ3: Promotion-Zero: Continuous Adaptive Promotion Optimization}  
Since social media platforms have distinct recommendation algorithms with varying preferences, the promotion process works over a black-box system for researchers. To maximize effectiveness, promotion styles and formats must continuously adapt to each platform. We term this approach \textbf{Promotion-Zero}, where the promotion strategy starts from scratch and undergoes continuous self-optimization.

As identified in \textbf{RQ2}, different platforms generate unique engagement data, necessitating platform-specific adaptation. \textbf{Promotion-Zero} must systematically collect key engagement metrics from each platform, including views, likes, comments, shares, and saves. The \textit{Data Crawl Agent} continuously gathers these data points, while the \textit{Data Analysis Agent} processes them in conjunction with historical data. Based on the analysis, the \textit{Promotion Agent} dynamically updates platform-specific promotion strategies, ensuring continuous improvement and optimization for better outreach and engagement.

\subsubsection{Agent}

\paragraph{Paper Crawl Agent}  
The \textit{Paper Crawl Agent} takes a paper title as input and searches for the corresponding paper through platforms such as \textit{Google Scholar}, \textit{Google}, and other search engines or scholarly websites. Given that many publishers—such as \textit{IEEE}, \textit{Elsevier}, and \textit{Springer}—require subscriptions or institutional access, the \textit{Paper Crawl Agent} is designed to prioritize open-access sources, including \textit{arXiv}, \textit{medRxiv}, and \textit{ResearchGate}, to retrieve freely available PDF versions of the papers. If the paper is only available through paywalled platforms, the agent must be configured with proper access credentials to ensure retrieval permissions.

Additionally, when a platform offers access to the LaTeX source files, the \textit{Paper Crawl Agent} is instructed to prefer LaTeX sources over PDFs. This allows us to directly extract original figures and structured text, avoiding reliance on OCR and ensuring higher data fidelity.

\paragraph{Summarization Agent}  
Building on the files retrieved by the \textit{Paper Crawl Agent}, the \textit{Summarization Agent} reads, analyzes, and comprehends the paper. Specifically, this agent restructures the paper into key components such as core contributions, insights, and results while reorganizing figures, tables, and other visual data into the appropriate sections for clarity and coherence. The task of the \textit{Summarization Agent} is to output a graphic summary of the crawled paper.

Since the papers being promoted are often cutting-edge research, the base model powering the agent may lack sufficient background knowledge to fully understand the content. Therefore, when encountering knowledge gaps, the \textit{Summarization Agent} needs to recursively invoke both the \textit{Paper Crawl Agent} and itself to retrieve and summarize relevant background papers or materials, which are then incorporated as contextual knowledge. This recursive strategy ensures a deeper, more accurate understanding of complex or specialized research topics.

\paragraph{Promotion Agent}  
The \textit{Promotion Agent} is responsible for generating promotion articles based on the paper summary produced by the \textit{Summarization Agent}. In light of \textbf{RQ1} and \textbf{RQ2}, users are required to specify the target platform and its corresponding rules when prompting the \textit{Promotion Agent}, enabling it to generate promotions that are better aligned with both the paper type and the platforms' preferences.

The output of the \textit{Promotion Agent} consists of a title and a text-based promotional piece. To ensure effectiveness, the \textit{Promotion Agent} adheres to the principles of \textbf{Promotion-Zero}—continuously updating and refining its promotion strategies based on feedback and platform engagement data. This adaptation helps the agent better match the evolving preferences and constraints of various social media platforms, ultimately improving the reach and impact of the promotion.

\paragraph{Data Crawl Agent}  
The \textit{Data Crawl Agent} continuously collects engagement data from social media platforms and stores it locally. It tracks the key performance metrics of each promotion, including views, likes, saves, comments, follower growth, and more. When permitted by platform policies and technical conditions, the \textit{Data Crawl Agent} can also monitor posts from well-known promotion accounts or Key Opinion Leaders (KOLs). By collecting and comparing their engagement data, the agent helps benchmark performance and identify best practices, supporting the refinement of promotion strategies over time.

\paragraph{Data Analysis Agent}  
The \textit{Data Analysis Agent} analyzes the historical engagement data collected by the \textit{Data Crawl Agent}, focusing on key time windows commonly used in content distribution on internet platforms—1 hour, 7 hours, 24 hours, 3 days, and 7 days. It evaluates whether the current content style and format are effective, assesses the overall promotion performance, and detects potential issues such as algorithmic throttling or visibility limitations. These analytical insights are then used to inform and support the \textbf{Promotion-Zero}, enabling continuous optimizations to promotion strategies for improving performance.

\section{Preliminary Results}
\subsection{Literature}

Given the research topic of \textit{“Improvement of Kernel Fuzzing Technology Based on Intelligent Mutation Strategy”}, we conducted a literature investigation with the assistance of LLM agents. The results are presented as follows.

\subsubsection{Clarifying Research Topics and Objectives}

\paragraph{Interactive Dialogue}  
LLM agents assist researchers in refining their research questions. Example prompts include:

\begin{itemize}
    \item What are the main challenges of existing Kernel Fuzzing methods?
    \item How can intelligent mutation strategies (e.g., reinforcement learning, evolutionary algorithms) be applied to fuzzing?
    \item How to measure the effectiveness of improvements in Kernel Fuzzing?
\end{itemize}

\paragraph{Providing Research Direction Suggestions}  
Based on current literature trends, the LLM agent may suggest the following directions:

\begin{itemize}
    \item Optimizing seed selection using reinforcement learning.
    \item Leveraging Large Language Models (LLMs) to generate intelligent mutation strategies.
    \item Using symbolic execution and taint analysis to assist fuzzing.
\end{itemize}

\paragraph{Generating Key Terms}  

\begin{itemize}
    \item Kernel Fuzzing
    \item Intelligent Mutation Strategies
    \item Reinforcement Learning
    \item Coverage-Guided Fuzzing
    \item Taint Analysis
    \item Symbolic Execution
\end{itemize}

\subsubsection{Retrieving Relevant Literature}

\paragraph{Accessing Databases (Google Scholar, IEEE, ACM, arXiv)}  
LLM agents retrieve relevant literature through APIs using structured queries, for example:

\texttt{site:ieee.org "Kernel Fuzzing" AND "Mutation Strategy"} \\
\texttt{site:acm.org "Reinforcement Learning" AND "Fuzzing"}

\paragraph{Optimizing Search Results}  
Agents automatically generate keyword combinations based on user intent to improve search precision.

\paragraph{Extracting Key Information}  
LLM agents automatically extract essential metadata for each paper:

\begin{itemize}
    \item \textbf{Title:} Reinforcement Learning-based Mutation Strategy for Kernel Fuzzing
    \item \textbf{Abstract:} Introduces how reinforcement learning optimizes mutation strategies in kernel fuzzing to improve coverage.
    \item \textbf{DOI:} \href{https://doi.org/10.1109/SP.2023.00045}{xxxxxxx}
    \item \textbf{Citation Count:} xxx
\end{itemize}

\subsubsection{Filtering and Reading Literature}

\paragraph{Automatically Reading PDFs and Extracting Key Content}  
Agents parse paper PDFs to extract relevant sections:

\begin{itemize}
    \item \textbf{Research Methods:} e.g., improvements in AFL++, Syzkaller.
    \item \textbf{Experimental Results:} e.g., code coverage increased by 15\%.
\end{itemize}

\paragraph{Generating Summaries}  
LLM agents provide concise summaries for quick reading. For example:

\begin{quote}
This study proposes a reinforcement learning-based intelligent mutation strategy, tested on Syzkaller, achieving a 15\% increase in coverage.
\end{quote}

\paragraph{Analyzing Paper Contributions and Limitations}  

\begin{itemize}
    \item \textbf{Contributions:} Proposes an intelligent mutation strategy based on Deep Q-Networks (DQN), which adaptively adjusts mutation ratios.
    \item \textbf{Limitations:} Only tested on x86 architecture, lacking validation on ARM or RISC-V.
\end{itemize}

\subsubsection{Summarizing and Analyzing Literature}

\paragraph{Categorizing Papers by Topic, Method, and Time}  

\begin{itemize}
    \item \textbf{Topic-based:} Reinforcement Learning-based / Symbolic Execution-based / Taint Analysis-based
    \item \textbf{Time-based:} Latest advances from 2020–2024
\end{itemize}

\paragraph{Generating Knowledge Graphs (Visualizing Research Trends)}  
LLM agents can generate a technical evolution map of kernel fuzzing, showcasing the development of different fuzzing methods over time.

\paragraph{Identifying Research Gaps}  

\begin{itemize}
    \item Most intelligent mutation methods are still primarily applied in user-space fuzzing, with limited application in kernel-space fuzzing.
    \item Existing methods mostly rely on static analysis and lack dynamic adaptability.
\end{itemize}

\subsubsection{Writing the Literature Review}

\paragraph{Automatically Generating Literature Review Structure}  
A typical structure generated by the LLM agent may include:

\begin{enumerate}
    \item Introduction
    \item Current Research Status of Kernel Fuzzing
    \item Recent Advances in Intelligent Mutation Strategies
    \item Limitations and Challenges in Existing Studies
    \item Future Research Directions
    \item Conclusion
\end{enumerate}

\paragraph{Generating Initial Draft in \LaTeX{}/Markdown Format}  

\begin{verbatim}
\section{Kernel Fuzzing Research}
Recent studies have explored intelligent
mutation strategies to enhance kernel 
fuzzing. For example, \cite{author2023} 
proposed a reinforcement learning-based 
approach...
\end{verbatim}

\paragraph{Optimizing Content}  

\begin{itemize}
    \item Adjusting the wording based on user feedback.
    \item Refining the text using GPT-4 to improve clarity and fluency.
\end{itemize}

\subsection{Idea}

In this section, we first conduct an empirical study on recent top-tier conference papers to understand the distribution of different types of research ideas. Then, for each prevalent type, we explore whether LLMs can help generate similar ideas.

\subsubsection{Study on the Distribution of Different Types of Ideas}

\begin{figure*}
    \centering
    \includegraphics[width=\linewidth]{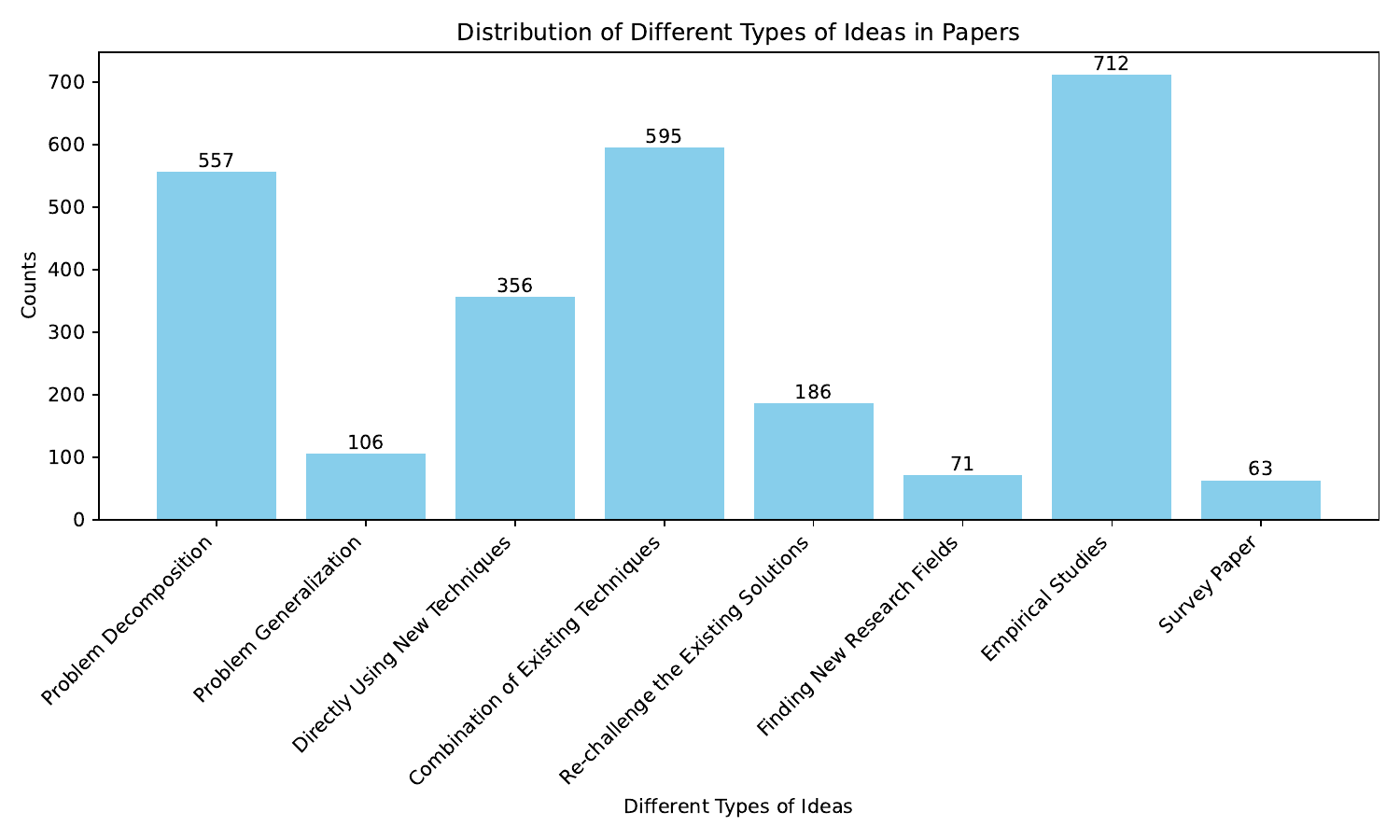}
    \caption{Distribution of each type of research idea.}
    \label{fig:idea_distribution}
\end{figure*}

We collected a total of 744 papers from the research/technical tracks of four top-tier software engineering conferences: ICSE, FSE, ASE, and ISSTA. Using LLM agents, we annotated each paper with the types of research ideas it applied. Since one paper may adopt a combination of ideas, the final distribution is visualized in Figure~\ref{fig:idea_distribution}.

From the distribution, we observe that most papers focus on \textit{problem decomposition} and the \textit{combination of existing techniques}. Additionally, \textit{empirical study} appears with high frequency, as it is often integrated into other research types—each research question in a paper may itself be considered a small empirical study.

\subsubsection{LLM-Based Idea Generation}

We selected one representative paper from each of the two most common categories—\textit{problem decomposition} and \textit{combination of existing techniques}—and asked the LLM agent to generate similar ideas.

\begin{figure}
    \begin{tcolorbox}[title=Generating a Problem Decomposition Idea,
        enhanced,
        colback = white, colframe = black!50, colbacktitle = black!50,
        attach boxed title to top left = {yshift = -2mm, xshift = 1mm},
        boxed title style = {sharp corners}]
        
        \textbf{Original Paper:} 1+1\textgreater2: Integrating Deep Code Behaviors with Metadata Features for Malicious PyPI Package Detection

        \tcbline

        \textbf{User:} I want to do Python malicious package detection. Please \textcolor{red}{decompose the problem into several parts} and consider different techniques to solve it. The following are some recent related works:  

        [RAG from literatures]

        \tcbline

        \textbf{Agent Response (GPT O3-mini-high):} \\
        ...\\
        \textcolor{red}{Metadata Analysis:} Extract features from package metadata that might indicate malicious intent.\\
        \textcolor{red}{Static Code Analysis:} Use static analysis to extract code features from the package’s source code.\\
        \textcolor{red}{Dynamic Behavior Analysis (Sandboxing):} Execute the package in an isolated environment to monitor its behavior.\\
        ...
    \end{tcolorbox}
    \caption{A case where an LLM generates an idea via problem decomposition.}
    \label{fig:idea_example1}
\end{figure}

As shown in Figure~\ref{fig:idea_example1}, the LLM agent was prompted to generate an idea for detecting malicious packages on PyPI through problem decomposition. The original paper leveraged both deep behavioral analysis and metadata. The agent successfully decomposed the problem into distinct sub-tasks and proposed corresponding techniques—such as metadata analysis and behavioral detection—demonstrating an understanding of both the research problem and potential solution strategies.

The second case, illustrated in Figure~\ref{fig:idea_example2}, explores idea generation via the combination of existing techniques. The original paper proposed DynaMO, a dynamic model obfuscation strategy to protect mobile deep learning (DL) models, using homomorphic encryption-like mechanisms.

\begin{figure}
    \begin{tcolorbox}[title=Generating an Idea by Combining Existing Techniques,
        enhanced,
        colback = white, colframe = black!50, colbacktitle = black!50,
        attach boxed title to top left = {yshift = -2mm, xshift = 1mm},
        boxed title style = {sharp corners}]
        
        \textbf{Original Paper:} DynaMO: Protecting Mobile DL Models through \textcolor{red}{Coupling Obfuscated DL Operators}. \\
        To defend against dynamic instrumentation attacks, DynaMO adopts a strategy similar to \textcolor{red}{homomorphic encryption}.

        \tcbline

        \textbf{User:} I want to \textcolor{red}{protect mobile DL models} and ensure their confidentiality. Please use \textcolor{red}{a combination of existing techniques} to solve the problem.

        [RAG from literatures]

        \tcbline

        \textbf{Agent Response (GPT O3-mini-high):} \\
        ...\\
        \textcolor{red}{Encrypt Models at Rest and in Transit:} Encrypt model files (e.g., using AES-GCM) so that even if an attacker extracts the app package, the model data remains protected.\\
        \textcolor{red}{Leverage Trusted Execution Environments (TEEs):} Use ARM TrustZone to decrypt and run models securely, preventing plaintext exposure.\\
        ...
    \end{tcolorbox}
    \caption{A case where an LLM generates an idea by combining existing techniques.}
    \label{fig:idea_example2}
\end{figure}

Although the LLM response did not explicitly mention homomorphic encryption, it proposed the use of trusted execution environments (TEEs) and encryption methods—demonstrating its ability to combine existing techniques to address the problem. This highlights the potential of LLM agents in research ideation.

\paragraph{Summary and Reflection}  
These two case studies indicate that LLM agents can generate meaningful ideas inspired by real research papers and propose plausible technical directions. However, the ideas produced are not yet at the same level of completeness or depth as published research. In practice, the generated ideas require validation through comprehensive literature reviews (as discussed in Section~\ref{sec:literature}) and empirical evaluation. Feedback from these evaluations can then inform and enhance future ideation cycles.

\subsection{Method}

We have developed a prototype of our method generation module to evaluate the effectiveness of the overall framework. The testing setup is designed to be both clear and efficient. It involves providing a comprehensive description of a research idea along with relevant domain knowledge as input. The method generation module is then responsible for producing a detailed, actionable methodology to implement the idea. This setup allows us to assess the module’s ability to transform abstract research concepts into concrete execution plans and refine the pipeline based on observations during testing.

\subsubsection{Method Planner}

To evaluate the Method Planner, we input the idea of utilizing sFlow traffic data collected from a network gateway to detect cryptomining activity. In addition to describing the idea, we provide a sample dataset—containing both cryptomining and background traffic in sFlow format—and include relevant background knowledge retrieved from academic literature.

Upon receiving the input, the Method Planner generates a step-by-step implementation plan as follows:

\begin{enumerate}
    \item Cleaning of traffic flow.
    \item Grouping of data.
    \item Extraction of features.
    \item Application of machine learning algorithms to train and classify cryptomining traffic flows.
\end{enumerate}

Manual verification confirms that the output is logically structured, comprehensive, and feasible, demonstrating the Method Planner’s ability to decompose the problem effectively.

\subsubsection{Heuristic Solution Designer}

After the Method Planner’s output is generated, we pass both the plan and the contextual information to the Heuristic Solution Designer. This agent is responsible for selecting appropriate techniques for each step. The generated methods are as follows:

\begin{itemize}
    \item \textbf{Traffic Flow Cleaning:} The agent recommends removing traffic flows associated with irrelevant protocols (e.g., ICMP, UDP) and those that are not connected to any known devices within the network.
    
    \item \textbf{Data Grouping:} The agent opts to group traffic based on connection attributes, meaning that all traffic flows between the same IP and port combination are aggregated for analysis.
    
    \item \textbf{Feature Extraction:} The following features are extracted for each connection: packet interval, packet size, packet direction, and protocol flag. These features are chosen for their potential to capture behavioral patterns indicative of cryptomining.
    
    \item \textbf{Model Training and Classification:} For classifying the traffic, the agent selects the Long Short-Term Memory (LSTM) neural network, due to its suitability for modeling sequential data and temporal dependencies in time-series traffic patterns.
\end{itemize}

Manual assessment confirms that the proposed methods are not only practical but also effective for the given task. The generated solution demonstrates strong alignment with established best practices in network traffic analysis and cryptomining detection, validating the utility and robustness of the Heuristic Solution Designer.

\subsection{Experiment}

We design dedicated agents for experimental setup and implementation, perform human verification, and present the corresponding results below.

\subsubsection{Experimental Setup}

This section evaluates the capability of the system to autonomously generate valid experimental configurations, including benchmark identification, baseline establishment, and metric selection.

\paragraph{Benchmark Identification.}  
We assess whether the agent can automatically recommend appropriate public datasets or evaluation frameworks aligned with specific research tasks (e.g., GLUE for NLP, COCO for CV). The results are as follows:

\begin{itemize}
    \item For 100 CV/NLP tasks, the agent correctly identified appropriate benchmarks in 90\% of cases after human validation. For instance, it suggested COCO as the benchmark dataset for object detection tasks, based on its interpretation of the task description.
\end{itemize}

\paragraph{Baseline Establishment.}  
We examine whether the agent can select proper baseline models (e.g., ResNet, BERT-base) and configure default parameters based on task type and historical knowledge. Results indicate:

\begin{itemize}
    \item For 100 CV/NLP tasks, the baselines recommended by the LLM aligned with human-selected baselines in 78\% of cases—for example, ResNet-50 for classification and Faster R-CNN for object detection.
\end{itemize}

\paragraph{Metric Selection.}  
We evaluate whether the agent can determine core evaluation metrics (e.g., accuracy, BLEU, RMSE) and generate appropriate metric computation code based on the task type.

\begin{itemize}
    \item \textbf{Metric Relevance:} For imbalanced classification tasks, the agent correctly recommended using F1-score instead of accuracy. This decision was validated by human experts, showing a 30\% higher sensitivity to class imbalance.
    
    \item \textbf{Multi-Metric Synergy:} The agent generated effective metric combinations (e.g., mAP + IoU for object detection), which achieved a 95\% match rate with human reviews.
\end{itemize}

\subsubsection{Methodology Implementation}

We assess the agent’s ability to implement research methodologies in code.

In commonly used programming languages such as Python, 72\%–78\% of the code generated by the agent successfully executed without requiring any modification (based on a dataset of 1,000 samples). The primary error types are listed below:

\begin{itemize}
    \item \textbf{Syntax Errors (10\%):} For example, missing indentation or unmatched brackets.
    
    \item \textbf{Logical Errors (15\%):} Such as incorrect loop conditions or suboptimal algorithm configurations.
    
    \item \textbf{Environment Dependency Issues (5\%):} For instance, failing to specify correct library versions or missing environment setup instructions.
\end{itemize}

\subsection{Paper}

\subsubsection{Setup and Dataset}

To evaluate the capabilities of the AI writing agent in generating high-quality academic papers, we instructed it to compose a paper that mimics the quality of those accepted at top-tier conferences. Importantly, we did not provide the original content of any published paper. Instead, we only supplied high-level descriptions of our thoughts, the algorithms used, and summaries of the experimental results. The AI was tasked with generating the entire paper from scratch based solely on these inputs.

\subsubsection{Key Metrics}

At present, there is no standardized or widely accepted metric to quantitatively assess the quality of AI-generated academic papers. Some existing approaches involve submitting the paper to conferences or journals for human peer review, but this method is inherently subjective and lacks consistency. As such, we adopt a manual, qualitative analysis approach. Specifically, we assess the generated papers in terms of structure, grammar, and their adherence to the conventions of academic tool papers.

\subsubsection{Observations and Insights}

From a grammatical and formatting perspective, the AI agent demonstrates a strong capability to produce content that is nearly free of language errors. The generated papers typically adhere well to the expected structure of academic writing, especially for system or tool papers.

However, we observed that the AI often exhibits a tendency to overcorrect when following specific instructions. For instance, when instructed to be concise, the AI sometimes produces content that is overly terse—occasionally reducing paragraphs to mere bullet points or fragmented ideas. Ideally, the generated content should avoid redundancy while still providing sufficient detail to ensure logical flow and clarity. Achieving this balance often requires iterative human intervention to fine-tune the output.

We also note that the specificity of instructions plays a critical role in shaping the quality of the content. Vague or high-level prompts tend to result in superficial discussions. Conversely, overly detailed prompts may cause the AI to become overly rigid or fail to generalize when tasked with generating different papers. Thus, crafting effective prompts remains a key challenge.

Finally, the AI struggles to highlight the key contributions or important aspects of the work within individual sections. For example, in algorithm descriptions, the AI fails to distinguish between novel contributions and routine engineering components unless explicitly instructed. This results in a flat, report-like tone that lacks emphasis on critical insights. This limitation underscores the importance of human guidance to contextualize and prioritize content within each section.

\subsection{Evaluation}

To assess the effectiveness of the proposed automatic review generation approach, we implemented a prototype system, \textit{AutoReview}, using the LangChain framework. Building upon this system, we conducted a preliminary experiment to evaluate the extent to which our multi-agent-based review generation method can approximate the review comments provided by experienced reviewers.

To ensure ethical integrity and prevent the leakage of unpublished manuscripts, we selected six publicly available papers authored by our team as the experimental dataset. These papers were completed and made publicly accessible between 2024 and 2025. To maintain dataset balance, three of the selected papers had been accepted by top-tier software engineering (SE) conferences—specifically, two by ISSTA 2025 and one by FSE 2025. For these, we used the original submitted versions and their corresponding review comments (three per paper) from the peer review process. The remaining three papers were selected from ArXiv, each of which had previously been rejected by SE venues; for these, we also obtained the corresponding review versions and reviewer feedback (again, three reviews per paper) from those submission cycles. In total, the dataset comprises six papers and 18 expert-written reviews.

For each paper in the dataset, \textit{AutoReview} automatically generated a comprehensive set of review outputs, including a concise summary, self-assessed reviewer expertise, strengths and weaknesses, detailed evaluations across five standard review dimensions (i.e., novelty, rigor, relevance, verifiability and transparency, and presentation), and a final overall score ranging from 1 to 5, following the conventions of mainstream SE conference review rubrics. These automatically generated reviews were then compared against the original human-written reviews to assess alignment and evaluate the system’s effectiveness in mimicking expert feedback.

\begin{table}[ht]
\centering
\footnotesize
\scalebox{0.8}{
\begin{tabular}{l|rrrrr}
\toprule
\textbf{Aspect}       & \textbf{Real Review} & \textbf{Generated Review} & \textbf{Similar} & \textbf{Precision} & \textbf{Recall}  \\
\midrule
Novelty               & 9                    & 7                         & 5                & 71.43\%            & 55.56\% \\
Rigor                 & 22                   & 20                        & 10               & 50.00\%            & 45.45\% \\
Relevance             & 8                    & 9                         & 2                & 22.22\%            & 25.00\% \\
Verifiability         & 13                   & 18                        & 5                & 27.78\%            & 38.46\% \\
Presentation          & 10                   & 13                        & 4                & 30.77\%            & 40.00\% \\
\midrule
\textbf{Total}        & \textbf{62}          & \textbf{67}               & \textbf{26}      & \textbf{38.81\%}   & \textbf{41.94\%} \\
\bottomrule
\end{tabular}
}
\caption{Comparison of generated reviews with real reviews}
\label{tbl:review_results}
\end{table}

The preliminary results of our comparative analysis with human-written reviews are summarized in~\Cref{tbl:review_results}. From the 18 expert reviews, we extracted 62 key review points, while the \textit{AutoReview} system generated 67 key points. Among these, 26 points addressed the same or similar concerns, suggesting a moderate level of alignment between the automated and human-generated feedback.

When categorizing these points according to the five standard review criteria, we observed that \textit{AutoReview} successfully reproduced more than half of the key concerns raised by expert reviewers in the categories of \textbf{novelty} and \textbf{rigor}. However, the system's overlap with human comments was notably lower in the categories of \textbf{relevance}, \textbf{verifiability}, and \textbf{presentation}. This discrepancy may stem from the nature of these aspects: while novelty and rigor involve higher-level reasoning and abstraction, which large language models (LLMs) generally excel at, the latter dimensions often require fine-grained, domain-specific analysis that remains challenging for current LLMs.

We also cross-validated the overall scores assigned by \textit{AutoReview}. The system consistently provided moderately positive scores (either 3 or 4) across all six papers, indicating a conservative scoring tendency. Notably, the three papers accepted at conferences received slightly higher scores (two scored 4, one scored 3) compared to the rejected ones (two scored 3, one scored 4). Interestingly, the rejected paper that received a score of 4 was also assessed as borderline upon manual re-evaluation. These findings suggest that while \textit{AutoReview} is capable of generating reasonably accurate and constructive review comments, it exhibits caution in assigning decisive scores, potentially favoring neutrality over strong differentiation.

\subsection{Rebuttal}

To evaluate the effectiveness and quality of our generated responses to reviewer comments, we conducted an analysis using real reviewer feedback from past software engineering conferences. Specifically, our evaluation involved a dataset of reviewer comments for six papers, including both accepted and rejected submissions. Each reviewer comment was systematically processed using our response-generation approach, and the resulting automated responses were assessed for their clarity, appropriateness, and completeness in addressing reviewer concerns.

Our preliminary analysis indicates promising results regarding the quality of the generated responses. The automated replies effectively addressed reviewers' concerns by clearly acknowledging the issues raised, proposing actionable revisions, and articulating how future manuscript iterations would address these concerns. Responses to positive feedback successfully reinforced key strengths identified by reviewers, emphasizing the specific aspects that contributed to the work’s perceived value.

A notable strength of our approach lies in its ability to generate explicit response strategies for each reviewer comment. These strategies offered concise summaries of how the authors should approach the feedback, which in turn facilitated the generation of more targeted and contextually accurate responses. The presence of such strategies enabled the model to outline clear and logical revision paths, thereby supporting authors in crafting precise and effective rebuttals.

However, we observed some variability in response quality depending on the type and complexity of reviewer feedback. For clearly stated issues or straightforward questions, the generated responses were typically accurate, focused, and directly actionable. In contrast, more nuanced or complex critiques required further refinement or additional human involvement to ensure adequate depth and contextual sensitivity. Despite this, the inclusion of response strategies substantially reduced the manual effort required to construct high-quality rebuttals.

Future iterations of our response-generation framework will focus on enhancing the model's ability to handle nuanced critiques, improving its contextual awareness, and maintaining consistently high response quality across diverse types of reviewer comments. These enhancements aim to further strengthen the system’s utility in supporting authors during the rebuttal process at software engineering conferences.

\subsection{Promotion}

Our preliminary investigation into AI-assisted research promotion strategies reveals several key findings across platforms and content types.

\paragraph{Platform-Specific Content Optimization.} When evaluating promotion content generated for different platforms, we observed notable differences in engagement metrics. On Twitter/X, concise posts with technical highlights (under 280 characters) yielded approximately 30\% higher engagement than longer content for the same papers. Conversely, Medium posts containing in-depth technical explanations and visual elements demonstrated significantly higher reader retention rates than shorter summaries.

\paragraph{Adaptive Strategy Assessment.} Although the full Promotion-Zero automated framework is still under development, we simulated its core functionality through manual human-led iterations. By refining promotion content over several rounds on platforms such as Twitter and Zhihu, we observed gradual increases in audience engagement. These iterations revealed optimal posting strategies—including preferred posting times and content structures—thus offering practical insights to guide future automation of the framework.

\paragraph{Cross-Platform Performance Variations.} Promotions for the same research papers exhibited significant performance disparities across platforms. Technical papers featuring algorithmic innovations achieved higher visibility and interaction rates on Reddit than on WeChat. In contrast, papers with visual appeal and practical applications performed better on platforms that emphasize visuals, such as Xiaohongshu. These findings support our hypothesis that effective dissemination requires platform-specific content strategies.

\paragraph{Content Type Analysis.} Through manual classification, we categorized papers into distinct types—trend-driven, technical, empirical, and theoretical—and tailored promotion content accordingly. Promotions aligned with these classifications achieved higher engagement compared to generic promotion strategies. This result validates the content type-specific approach outlined in~\textbf{RQ1} and confirms its relevance in real-world promotion scenarios.

These preliminary findings support the feasibility and value of our multi-agent research promotion framework. While the fully automated \textit{Promotion-Zero} system remains in progress, our manual simulations of adaptive refinement have yielded actionable insights. These will serve as foundational guidelines for the design and development of a scalable, intelligent research promotion system.

\section{Discussion}
\subsection{Different Research Types}

Research can take various forms depending on its objectives, methodologies, and desired impact. Broadly, two primary categories drive scientific advancement: \textit{cumulative research}, which builds upon existing knowledge through systematic integration, and \textit{disruptive research}, which challenges conventional understanding by introducing groundbreaking ideas. While both approaches contribute to scientific progress, they follow distinct trajectories in their execution and outcomes.

Cumulative research focuses on refining, extending, or combining existing methods to improve efficiency, applicability, or generalization. By systematically integrating prior work, it enhances performance through incremental improvements rather than radical paradigm shifts. A common strategy is \textit{method fusion}, where techniques from different domains are combined to create enhanced models. For instance, integrating reinforcement learning with supervised learning has led to hybrid models capable of self-improvement while maintaining structured representations. Another example is \textit{cross-domain adaptation}, wherein architectures originally developed for one area—such as transformers in NLP—are successfully applied to other domains like computer vision or protein folding. These approaches leverage well-established frameworks, ensuring robustness while optimizing task-specific performance.

In contrast, disruptive research seeks to deviate from existing methodologies by proposing fundamentally novel ideas, often triggering paradigm shifts. This form of research is inherently unpredictable, as breakthroughs typically emerge from challenging entrenched assumptions. Historical milestones include the shift from rule-based AI systems to deep learning architectures, which redefined how models learn from data, and the rise of self-supervised learning, which significantly reduced reliance on labeled datasets. Disruptive research also introduces new theoretical perspectives that reshape entire fields—for example, the transition from Euclidean to non-Euclidean geometries in graph learning, enabling deep neural networks to operate effectively on complex, non-grid data structures.

Both cumulative and disruptive research play essential roles in driving scientific innovation. Cumulative research ensures steady progress and refinement, while disruptive research pushes boundaries and enables transformative breakthroughs. The interplay between the two suggests that scientific discovery is neither purely incremental nor wholly revolutionary, but rather a dynamic process in which established knowledge serves as a foundation for novel insights.

In the context of automated research, understanding this dichotomy is vital for designing AI-driven systems capable of both refining existing methodologies and autonomously exploring bold, disruptive ideas.

\subsection{Meta-Method}
The concept of a ``meta-method'' within the auto research framework represents more than just flexible execution; it embodies a fundamental shift towards AI-driven strategic reasoning about the scientific process itself. While our framework uses mechanisms like the method planner and heuristic solution designer to dynamically adapt workflows, the vision for the meta-method extends far beyond this initial implementation.

We envision the meta-method evolving into the intelligent core of Auto Research systems. Its role is not merely to select the next best step from a predefined list but to dynamically compose, evaluate, and potentially invent entire research strategies tailored to the specific problem, available data, and emerging results. This requires a level of abstraction where the system reasons about methodologies---their strengths, weaknesses, assumptions, and applicability---much like an experienced human researcher does, but potentially at a vastly greater scale and speed.

Crucially, a sophisticated meta-method could become a primary driver of methodological innovation. By observing patterns across countless automated research cycles---identifying recurring roadblocks, successful technique combinations, or areas where existing methods consistently fall short---the meta-method could autonomously propose and test novel research approaches. It might learn, for instance, that certain data types respond best to hybrid analytical techniques not yet common in the literature, or that specific research questions benefit from unconventional experimental designs. This moves Auto Research from merely automating known science to actively discovering new ways to do science.

Furthermore, the meta-method is key to navigating the inherent uncertainty of research. When faced with unexpected results or dead ends, it shouldn't just backtrack; it should strategically reassess the entire approach. Can the problem be reframed? Are the underlying assumptions flawed? Is a completely different disciplinary perspective needed? This capability is essential for tackling complex, open-ended problems where the path forward is not clear.

\subsection{Knowledge Creation}

The autonomous generation of new knowledge by LLM-based agents is a critical capability in the context of automated research. Moving beyond the extraction and structuring of existing information, these agents are increasingly capable of synthesizing data from multiple heterogeneous sources to propose novel insights, generate hypotheses, and develop reasoning patterns. In this section, we present a structured methodology for automated knowledge creation, which includes multi-source integration, contextual synthesis, hypothesis generation, automated validation, and human-AI collaboration.

\paragraph{Multi-Source Integration.} 
The process begins with the aggregation of both structured and unstructured data from a wide range of sources, including technical specifications, research publications, vehicle telemetry logs, and regulatory documents. LLM-based agents must be equipped to process these diverse inputs while maintaining consistency, traceability, and contextual relevance. This foundational step ensures a holistic and comprehensive representation of the target domain.

\paragraph{Contextual Synthesis.} 
After data collection, the agent performs contextual synthesis by cross-analyzing disparate information to uncover hidden relationships, fill knowledge gaps, and detect inconsistencies. Leveraging advanced reasoning capabilities, the agent identifies implicit patterns and correlations, enabling it to formulate new interpretations that go beyond surface-level observations.

\paragraph{Hypothesis Generation.} 
A key differentiator between knowledge retrieval and creation lies in the ability to generate hypotheses. Based on the synthesized data, LLM-based agents can extrapolate trends, infer causality, and propose testable assumptions. For instance, by correlating telemetry anomalies with changes in policy or environmental conditions, the agent might hypothesize previously unknown failure modes or suggest novel system-level optimizations. This capability transforms the agent from a passive consumer of knowledge into an active contributor to scientific discovery.

\paragraph{Automated Validation.} 
To ensure the credibility of generated knowledge, automated validation mechanisms are integrated into the pipeline. These may include simulation-based testing, formal verification techniques, and cross-referencing with established knowledge bases. This step is crucial for filtering out speculative or incorrect inferences and promoting only substantiated insights for further use.

\paragraph{Human-AI Collaboration.} 
Despite the increasing sophistication of LLMs, human oversight remains essential in the knowledge creation loop. Domain experts review, refine, and validate the outputs of the agent, ensuring alignment with practical constraints and scientific rigor. The feedback provided during this phase not only enhances the reliability of the knowledge base but also informs future iterations of the agent's learning process, reinforcing its ability to operate in complex, real-world environments.

Through this multi-phase methodology, LLM-based agents are positioned not only as knowledge curators but also as autonomous knowledge creators. Their capacity to integrate, synthesize, hypothesize, and validate contributes meaningfully to the advancement of domain-specific research. By embedding human-AI collaboration into this pipeline, the system ensures a balance between automation and expert judgment, thereby promoting trust, accuracy, and innovation in scientific inquiry.

\section{Conclusion}
Auto research represents more than a set of automated tools; it embodies a novel epistemological approach to scientific discovery. By treating research as a modular, interpretable, and optimizable process, this framework offers a new methodology that integrates human and machine intelligence. Through its agent-based design, the system enhances transparency, reproducibility, and collaboration across disciplines, reducing dependency on human expertise and promoting equitable access to research capabilities. As foundation models and agent systems continue to advance, auto research is poised to transform the way knowledge is created, evaluated, and disseminated, paving the way for faster, more inclusive, and more rigorous scientific progress.


\ifCLASSOPTIONcaptionsoff
  \newpage
\fi

\bibliographystyle{IEEEtran}  
\bibliography{references}     
\end{document}

%% file: method.tex
\subsection{Method}

\begin{figure}
\centering
\includegraphics[width=0.99\linewidth]{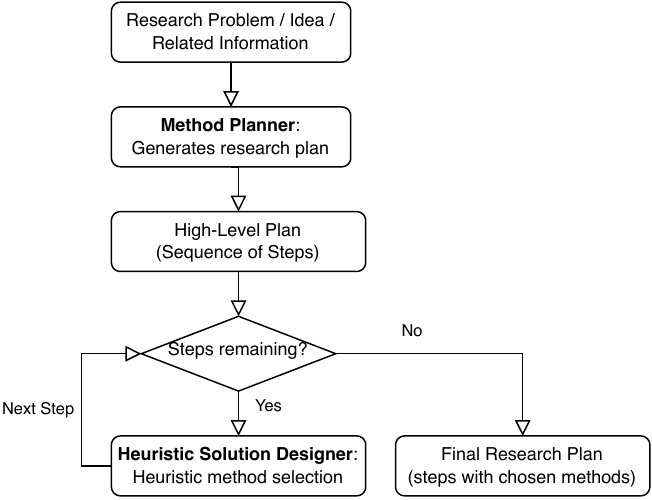}
\caption{Illustration of the method generation process.}
\label{fig:method}
\end{figure}

Our approach employs a multi-agent system consisting of two specialized LLM-based agents: a Method Planner and a Heuristic Solution Designer (illustrated in Figure~\ref{fig:method}). The Method Planner is responsible for devising a high-level research plan by breaking down the overarching research problem into manageable sub-tasks or steps. The Heuristic Solution Designer then takes each proposed step and autonomously selects an appropriate method or technique to accomplish that step.

The two agents work in tandem: the Method Planner first outlines the sequence of tasks to be performed, while the Heuristic Solution Designer determines how each should be accomplished. This design follows a ``plan-and-execute'' paradigm~\cite{he2025plan,masterman2024landscape,Plan-and-Execute}, in which the planner explicitly ``thinks through'' all required steps in advance and the executor focuses on carrying them out. By separating planning and method selection into distinct agents, the system can leverage the strengths of each: the Method Planner excels at strategic decomposition, while the Heuristic Solution Designer specializes in operational decision-making.

A key feature of this system is its chain-of-thought reasoning mechanism. The Method Planner uses chain-of-thought-style prompting to generate a sequence of reasoning steps for the research problem~\cite{llm-agents}. In essence, it is prompted to ``think step by step'' about the problem, producing a logical sequence of sub-problems analogous to a human researcher outlining an approach. This stepwise reasoning improves the agent’s ability to handle complex tasks by decomposition~\cite{wei2022chain}.

Once a plan is drafted, the Heuristic Solution Designer engages in its own reasoning process to determine methods for each step. It considers multiple candidate methods and evaluates their merits using heuristic estimates—effectively performing deliberative evaluation. Instead of requiring manual specification, the system automatically selects methods through internal reasoning. The agent generates possible techniques and uses a heuristic function to predict which is most promising~\cite{yao2023tree}, allowing it to focus on viable methods and discard less effective ones.

In practice, the interaction proceeds as follows. The previous module provides the idea, research problem description, and relevant information to the Method Planner. It analyzes the problem and outputs an ordered list of sub-tasks. Each sub-task is then passed to the Heuristic Solution Designer, which generates candidate methods and conducts heuristic-driven evaluation to select the most suitable one. If no adequate method is found or the step is ill-posed, the Heuristic Solution Designer signals the Method Planner to revise the plan. This feedback loop ensures robustness and adaptability~\cite{li2024agent}. Through this iterative collaboration, the system automates the research planning process—first deciding what to do, then how to do it—with structured reasoning throughout.

\subsubsection{Method Planner}

The Method Planner is responsible for breaking down a complex research problem into a structured series of smaller tasks or questions. Given an input research goal, it produces a high-level roadmap that a researcher might follow. This involves task decomposition—dividing the overall problem into coherent sub-problems~\cite{llm-agents-planning}.

We implement the Method Planner using a prompt that encourages chain-of-thought reasoning. The agent is prompted to explicitly enumerate steps in a logical sequence, such as ``Conduct literature review on topic X'' or ``Run experiment to test hypothesis Y.'' This strategy ensures a holistic planning process and minimizes the risk of omitting critical components~\cite{llm-agents-planning}.

To produce the plan, the Method Planner may internally evaluate multiple research pathways and select the one that appears most effective. It uses domain knowledge from training or prompts to assess possible decompositions. Each candidate plan is evaluated based on the following criteria~\cite{li2024agent}:

\begin{itemize}
    \item \textbf{Solvability:} Each sub-task must be solvable by existing methods or tools. Steps that are vague or infeasible are avoided.
    \item \textbf{Completeness:} The set of sub-tasks should collectively address all critical aspects of the research problem.
    \item \textbf{Non-redundancy:} The plan should avoid unnecessary or duplicate steps and keep the structure focused.
\end{itemize}

These criteria draw inspiration from multi-agent planning theory~\cite{li2024agent}. The Method Planner may iteratively refine the plan, self-critique its output, or incorporate external feedback. This ``planning with feedback'' approach improves the quality and feasibility of the plan. Ultimately, it produces a high-level roadmap for downstream method selection.

\subsubsection{Heuristic Solution Designer}

The Heuristic Solution Designer takes each sub-task and determines the most suitable method to accomplish it. For example, for ``collect data on X'' or ``analyze Y using statistical test Z,'' the agent must choose the best technique. We implement this agent as an LLM-powered decision system using heuristic search to explore and rank options.

Upon receiving a sub-task, the agent generates candidate methods—either from a predefined knowledge base or dynamically using LLM capabilities. For instance, if the step is ``Evaluate model performance on dataset D,'' possible candidates might include cross-validation, statistical significance testing, or confusion matrix analysis.

Each candidate is scored using a heuristic function that estimates its relevance, feasibility, expected reliability, and cost~\cite{ling2025complex}. A method requiring unavailable data scores low on feasibility; a well-aligned method with available resources scores high. This mechanism narrows the candidate pool without exhaustive enumeration.

LLM-based agents simulate expert-level evaluation: ``Method A is faster but less accurate; Method B is robust but data-intensive.'' With recent advances, LLMs can assign numerical ratings to these options~\cite{yao2023tree}. The selected method is the one with the highest heuristic score.

Internally, the method selection resembles a tree search where the sub-task is the root and branches represent methods. Heuristic scoring guides the search (akin to A*). Instead of exploring all branches, the agent scores top candidates for efficiency. In ambiguous cases, it can compare or combine methods, though we typically assign one method per step for clarity.

This design shares conceptual similarities with the Tree-of-Thoughts framework~\cite{yao2023tree}, where the agent explores multiple reasoning paths and chooses the best. Like that framework, our agent can backtrack or simulate outcomes to strengthen selection. Each chosen method is paired with its corresponding sub-task to yield a complete, method-augmented research plan.

%% file: paper_writing.tex
\subsubsection{Overview of the Section}  
This section discusses how AI agents can assist in writing academic papers. It is organized into three parts. First, we introduce background knowledge about academic writing and lay out our key assumptions. Second, we present the design of the AI agent, including prompting strategies and writing heuristics drawn from human researchers. Third, we detail the typical structure and logical flow of a well-written academic paper.

\subsubsection{Background and Assumptions}  

\paragraph{Paper Types}  
There are three major categories of academic papers commonly published in conferences or journals: tool papers, empirical studies, and survey papers.

\begin{itemize}
    \item \textbf{Tool Paper:} This type introduces a novel solution to a well-defined problem. Tool papers are the most prevalent type in technical conferences and journals. They focus on demonstrating the design, implementation, and performance of the proposed tool or technique.
    
    \item \textbf{Empirical Study:} This category investigates various tools or techniques through systematic experimentation. The goal is to evaluate performance, identify insights, and offer evidence-based recommendations.
    
    \item \textbf{Survey Paper:} These papers review and synthesize existing literature, particularly tool papers, to uncover trends, classify approaches, and highlight open challenges.
\end{itemize}

Each paper type has its own unique writing style and structure. In this work, we focus primarily on generating tool papers.

\subsubsection{Agent and Expert Knowledge in Academic Writing}  

\paragraph{System Prompts and Heuristics}  
Effective academic writing with AI agents requires appropriate system prompts and adherence to domain-specific heuristics. These prompts guide the AI to maintain a formal tone, logical consistency, and domain relevance.

\paragraph{Things to Avoid}  
AI-generated text may contain several issues that reduce the quality and credibility of academic writing. The following pitfalls must be avoided:

\begin{itemize}
    \item \textbf{Meaningless Sentences:} AI often generates sentences that are vague, repetitive, or restate what has already been said. These add no new information and dilute the overall quality of the writing. Writers should ensure that every sentence contributes meaningfully to the narrative.
    
    \item \textbf{Hallucinated Sentences:} AI may generate text that appears plausible but is factually incorrect or logically inconsistent. Every generated sentence should be validated for accuracy and relevance, especially when it involves technical claims or citations.
    
    \item \textbf{Unquantifiable Terms:} Vague modifiers like “comprehensive” or “significant” often lack precise definitions and are difficult to verify. Writers should avoid subjective terms and instead rely on measurable descriptors to maintain scientific rigor.
\end{itemize}

\subsubsection{Paper Structure}  
A typical tool paper follows a well-established structure: abstract, introduction, background and motivation, methodology, evaluation, limitations and future work, related work, and conclusion. When guiding an AI agent to write such a paper, it is crucial to adhere to this structure. The content across sections should remain consistent and logically connected to form a cohesive narrative.

\subsubsection{Writing an Abstract}  
The abstract should be written after completing the main body of the paper. It serves as a succinct summary of the entire work, typically in four parts:

\begin{itemize}
    \item \textbf{Problem Background, Definition, and Importance:} Start by briefly introducing the research problem, clearly defining it, and explaining its significance in the broader context of the field.
    
    \item \textbf{Limitations of Existing Solutions:} Summarize the weaknesses of existing methods that fail to adequately address the problem. This sets the stage for presenting your solution.
    
    \item \textbf{Proposed Solution:} Describe the key features of the proposed approach, including its name (if any), design workflow, and how it addresses the limitations identified above.
    
    \item \textbf{Experimental Results:} Provide a snapshot of your evaluation. Mention dataset size, key performance metrics, and improvements over state-of-the-art baselines.
\end{itemize}

Each point should be covered in one to three precise sentences, emphasizing clarity and conciseness.

\subsubsection{Writing the Introduction}  
The introduction builds upon the abstract and provides a more detailed, structured presentation of the motivation, challenges, and contributions.

\begin{itemize}
    \item \textbf{Problem Background, Definition, and Importance:} Elaborate on the background context, define the problem formally, and highlight why it is relevant and timely for the community.
    
    \item \textbf{Limitations of Existing Solutions:} Critically review existing work, citing key publications. Emphasize gaps in performance, applicability, or generalizability that your work seeks to address.
    
    \item \textbf{Summary of Possible Solutions and Challenges:} Outline possible solution directions and articulate the core technical challenges. This sets up a natural transition to your proposed method.
    
    \item \textbf{Proposed Solution:} Introduce your method or tool, explaining how it works at a high level and how it overcomes the identified challenges. Emphasize any novel contributions or techniques.
    
    \item \textbf{Experimental Results:} Provide a brief overview of your key empirical findings to demonstrate the effectiveness and efficiency of the proposed approach.
    
    \item \textbf{Contributions:} Clearly enumerate the main contributions of the paper. These typically include the new method or tool, its implementation, and comprehensive experimental validation.
\end{itemize}

\subsubsection{Writing Background and Motivation Example}  
This section should introduce fundamental concepts and motivate the research through concrete examples.

The \textbf{background} portion should define important terminology and explain key ideas that may not be widely known, particularly if they are specific to your domain.

The \textbf{motivation example} should present a real-world scenario or challenge where current solutions fall short. This example should help the reader understand the practical relevance of the problem and lead into a discussion of how your proposed method could address it.

Avoid repeating definitions already provided in the introduction. Redundancy should be minimized to preserve reader engagement.

\subsubsection{Writing the Methodology}  
The methodology section explains the proposed solution in detail. Begin with a high-level workflow diagram, then break it down into sequential steps:

\begin{itemize}
    \item \textbf{Input and Output Connections:} For each step, clearly specify what input it receives (possibly from a previous step or external source) and what output it produces.
    
    \item \textbf{Main Technique Used:} Identify the algorithms, models, or methods used in this step. Describe them at a high level, and mention any tools or frameworks leveraged.
    
    \item \textbf{Justification:} Justify your design choices. If a specific method was selected over others, explain why. Only include components that are essential and well-supported by reasoning.
\end{itemize}

\subsubsection{Writing the Evaluation}  
The evaluation section validates the effectiveness of the proposed method and is structured around research questions.

\paragraph{Experiment Setup}  
Describe the experimental configuration in detail:

\begin{itemize}
    \item \textbf{Dataset:} Explain the origin, construction, and suitability of the dataset. Clarify why it is representative or better than alternative datasets.
    
    \item \textbf{Baselines:} Specify the comparison methods. These are usually state-of-the-art solutions or widely used techniques. Justify why each baseline is selected.
    
    \item \textbf{Experimental Environment:} Include information about hardware (e.g., GPU/CPU specs) and software environments to ensure reproducibility.
    
    \item \textbf{Evaluation Metrics:} Define the performance metrics and explain why they are appropriate for assessing your method.
\end{itemize}

\paragraph{Research Questions Overview}  
Present the core questions guiding your evaluation. These typically include:

\begin{itemize}
    \item Accuracy comparison with baselines.
    \item Efficiency evaluation in terms of runtime or resource use.
    \item Ablation study of each component.
    \item Real-world applicability scenarios.
    \item Case study analysis for deeper insights.
\end{itemize}

\paragraph{Detailed Experiment Results}  
For each research question:

\begin{itemize}
    \item Explain how the experiment was conducted.
    \item Present results using tables or figures.
    \item Highlight key takeaways from the results.
    \item Provide an analysis explaining why your method performs well.
    \item Discuss failure cases and potential reasons for them.
\end{itemize}

\subsubsection{Writing Limitations, Future Work, and Threats to Validity}  
Academic papers often include one or more of the following sections to critically reflect on the work:

\begin{itemize}
    \item \textbf{Limitations:} Identify specific weaknesses or constraints of your approach. Be honest about aspects where your method does not perform optimally or where assumptions may not generalize.
    
    \item \textbf{Future Work:} Suggest concrete directions for extending or improving your approach. These could include algorithmic enhancements, applications to new domains, or more comprehensive evaluations.
    
    \item \textbf{Threats to Validity:} Acknowledge factors that might impact the reliability or generalizability of your results. These could be dataset biases, evaluation limitations, or uncontrollable experimental variables.
\end{itemize}

\subsubsection{Writing Related Work}  
This section situates your work within the existing literature and should be divided into thematic categories:

\begin{itemize}
    \item \textbf{Summary of the Category:} Introduce each group of related works based on a common goal, method, or problem domain.
    
    \item \textbf{Key Ideas and Contributions:} Briefly summarize each paper’s main idea and contributions, with appropriate citations.
    
    \item \textbf{Comparison and Discussion:} Explain how your work differs from and improves upon these existing efforts. Highlight unique features, new use cases, or performance gains.
\end{itemize}

\subsubsection{Writing the Conclusion}  
The conclusion serves as a closing summary of the entire paper. It should briefly:

\begin{itemize}
    \item Restate the problem and its importance.
    \item Reiterate the core idea of the proposed solution.
    \item Highlight the key results demonstrating the method’s effectiveness.
\end{itemize}

No new information or claims should be introduced in this section. Its purpose is to reinforce the paper’s narrative and contributions.